\documentclass{article}

\usepackage[preprint]{corl_2026} 

\usepackage{graphicx}
\usepackage{subcaption}
\usepackage{wrapfig}
\usepackage{booktabs}
\usepackage{multirow}
\usepackage{tabularx}
\usepackage{amsmath}
\usepackage{amssymb}
\usepackage{algorithm}
\usepackage{algpseudocode}
\usepackage[table]{xcolor}
\usepackage{url}
\usepackage{float}
\usepackage{soul}

\newcommand{\hlchair}[1]{\colorbox{pink!25}{#1}}
\newcommand{\hlcabinet}[1]{\colorbox{yellow!30}{#1}}
\newcommand{\hldoor}[1]{\colorbox{cyan!20}{#1}}
\newcommand{\hlmonitor}[1]{\colorbox{blue!15}{#1}}
\newcommand{\hlbench}[1]{\colorbox{green!18}{#1}}

\newcommand{\method}{VLM-GLoc}
\newcommand{\pose}{\mathbf{x}}

\newcommand{\emb}{\mathbf{e}}
\newcommand{\map}{\mathcal{M}}

\newcommand{\figplaceholder}[1]{%
  \fbox{\parbox[c][1.35in][c]{0.90\linewidth}{\centering Missing figure: \texttt{\detokenize{#1}}}}%
}
\newcommand{\safegraphics}[2][]{%
  \IfFileExists{#2}{\includegraphics[#1]{#2}}{\figplaceholder{#2}}%
}

\title{\method{}: Vision-Language Model Enhanced Monte Carlo Localization for Robust Semantic Global Localization in Cluttered Quasi-Static Environments}
\author{
  Shivendra Agrawal \quad Bradley Hayes \\
  University of Colorado Boulder \\
  Boulder, Colorado, USA \\
  \texttt{\{shivendra.agrawal, bradley.hayes\}@colorado.edu}
}

\begin{document}
\maketitle

\begin{abstract}
Global localization in geometrically aliased, quasi-static environments such as grocery stores, offices, schools, and hospitals poses a significant challenge for mobile robots. Grocery stores with parallel aisles and a long-tailed distribution of products, as well as offices and labs with repetitive furniture such as chairs, desks, monitors, and doors, exemplify common indoor environments that present geometric and even semantic ambiguity. Traditional approaches rely either on distinct geometric features or on domain-specific vision pipelines that struggle with long-tail semantic distributions and transient visual clutter. We present \method, a method for hierarchical semantic Monte Carlo Localization (MCL) that leverages open-vocabulary Vision-Language Models (VLMs) as a unified semantic observation front-end. We hypothesize a three-fold benefit from VLMs: (1)~extracting highly discriminative rich text features, (2)~implicit quality filtering of blurry or dynamic objects, and (3)~permanence reasoning for targeted data augmentation. We introduce an inverse semantic proposal mechanism that seeds particles via text-to-map retrieval. Evaluated across two real-world environments with different characteristics and two different platforms: a 3,500 sq. ft. grocery store with a cellphone and a 3,700 sq. ft. lab space with a quadruped, \method{} achieves 70\% and 74\% global localization success respectively, substantially outperforming traditional geometry-only and domain-specific baselines.

\end{abstract}



\section{Introduction}
\label{sec:intro}

Global localization (recovering a robot's pose within a known map) is a prerequisite for true autonomy. While classical Adaptive Monte Carlo Localization (AMCL) is the \emph{de facto} standard in robotics, it frequently fails in \emph{geometrically aliased} environments. In spaces like grocery store aisles, office corridors, or repetitive laboratory cubicles, the local occupancy structure is nearly identical across different global regions. 
Semantic information offers a natural solution: humans distinguish locations not by wall geometry, but by \emph{what they see}. However, existing semantic localization systems \cite{zimmermanLongTermLocalizationUsing2023,agrawal2026shelfaware} typically rely on top-level class vectors requiring domain-specific object detectors. This proves brittle against long-tail distributions. In repetitive offices and labs, a generic ``chair'', ``monitor'', or ``cabinet'' label lacks the granularity needed for disambiguation. 

We present \textit{\method}, a hierarchical Monte Carlo Localization framework that replaces rigid visual pipelines with a unified Vision-Language Model (VLM) interface. Rather than asking the VLM to directly regress metric coordinates, \method{} uses it as a rich semantic observation generator. These natural-language observations are embedded by a lightweight sentence encoder and compared against a pre-built semantic map via cosine similarity. 
Our central hypothesis is that VLMs provide a critical three-fold advantage for probabilistic localization:
\textit{1. Rich Text Features:} Moving beyond simple class labels to highly discriminative descriptions (e.g., from ``cabinet'' to ``yellow metal flammables cabinet'').
\textit{2. Implicit Quality Filtering:} The ability to label and suppress noisy detections, transient objects (e.g., jacket, cup), and dynamic actors (e.g., people, cart).
\textit{3. Permanence Reasoning \& Augmentation:} Classifying items as static, quasi-static, or dynamic, which enables targeted training data augmentation (semantic dropout) without physically altering the scene.

To exploit these advantages without suffering premature geometric collapse, \method{} employs a hierarchical tracking strategy. We introduce an \emph{inverse semantic proposal} mechanism where a single query image is inverted into a sparse set of likely pose hypotheses to seed the particle filter. Semantics are used first to narrow down the global search space, while geometry (lidar/depth) is utilized as a consistency signal to lock in precision once spatial convergence is detected.

\noindent \textbf{The contributions of this work are:} (1) A VLM-based semantic MCL method that uses open-vocabulary text embeddings as an observation model rather than for direct pose regression; and (2) An inverse semantic proposal mechanism and hierarchical fusion strategy to resolve geometric aliasing. Both are validated within two real-world domains: a 3,500 sq. ft. grocery store and a 3,700 sq. ft. laboratory, demonstrating that a single language interface substantially outperforms both geometry-only and domain-specific baselines.

\section{Related Work}
\textbf{Probabilistic and geometric global localization.}
Particle-filter localization remains a standard formulation for start-anytime
robot pose estimation because it represents multi-modal beliefs and naturally
fuses motion and sensor likelihoods. KLD-sampling and AMCL made this family of
methods practical for robot navigation \citep{foxKLDSamplingAdaptiveParticle2001,nav2_amcl}.
Depth-camera and lidar localization can be accurate when the environment has
distinctive geometry \citep{biswasDepthCameraBased2012,watanabeRobustLocalizationArchitectural2020},
but recent surveys emphasize that global lidar localization still struggles
under perceptual aliasing, dynamics, and map change \citep{yinSurveyGlobalLiDAR2024}.
Learned implicit or area-graph representations address some of these challenges
\citep{kuangIRMCLImplicitRepresentationBased2023,xieRobustLifelongIndoor2024},
but they still depend primarily on geometric structure. \method{} keeps the
probabilistic strengths of MCL while adding a language-rich observation channel
for places where geometry alone admits many plausible modes.

\textbf{Semantic and long-term localization.}
Semantic localization augments geometry with object, text, or layout cues.
Semantic SLAM and floor-plan alignment methods use objects and room structure to
localize in prior indoor maps \citep{goswamiEfficientRealTimeLocalization2023a,zimmermanLongTermLocalizationUsing2023}.
Other work exploits text spotting and named landmarks for robust onboard
localization \citep{zimmerman2022robust}, or models object
existence probabilistically to support long-term localization under change
\citep{adkins2022probabilistic}. These approaches show that semantic information can break geometric symmetries, but their representations are often fixed to a detector vocabulary or a specific landmark type. \method{} instead uses natural-language scene descriptions, preserving attributes such as color, material, readable text, packaging, and object state.


\textbf{Retail and mobile localization.}
Grocery stores pose unique localization challenges due to repetitive shelving geometry and a massive, dynamic semantic space. While prior work \citep{agrawal2026shelfaware} demonstrates that fixed-class shelf semantics support mobile localization, relying on predefined category vectors inherently limits the representation. \method{} expands this paradigm by fusing open-vocabulary product descriptions with a generalized particle filter. By moving beyond fixed ontologies, our approach capitalizes on the highly discriminative value of specific product names, brands, and packaging to resolve global ambiguity.


\textbf{Open-vocabulary maps and learned visual localization.}
Recent foundation-model systems build open-vocabulary 3D scene graphs for perception and planning \citep{gu2024conceptgraphs}.
Vision-and-language navigation studies language-conditioned movement and route following \citep{anderson2018vln,krantz2020vlnce,chen2024mapgpt}, but generally assumes known localization rather than solving it within a fixed prior map. Recent work pushes toward direct localization, utilizing VLMs for 3D reasoning and point-cloud grounding \citep{kang2026vlm,qu2026loc3rvlm}. These methods show that language-conditioned foundation models can support spatial grounding. \method{} takes a complementary robotics view: the VLM never outputs coordinates. It produces semantic evidence for a probabilistic filter, while multi-modal belief maintenance, odometry fusion, hierarchical lidar consistency, and recovery from dynamic or quasi-static scene changes remain inside MCL.

\section{Problem Formulation}

We assume a prior 2D occupancy map. First, \method{} builds an offline semantic map. At test time, the robot receives an RGB-D observation, an odometry increment, and optionally lidar measurements. The goal is global localization: estimate the robot pose \(\pose_t = (x_t, y_t, \theta_t)\) in the prior map, without assuming a known initial pose. Let \(\map = \{(p_i, \emb_i)\}_{i=1}^{N}\) be a semantic map over discrete pose cells \(p_i\). Each \(\emb_i\) is the normalized text embedding of the VLM description expected from that pose. Given a query image \(I_t\), a VLM front end produces a structured text observation \(T_t\), and a text encoder produces a normalized query embedding \(\emb_t\). The semantic likelihood for a particle \(x\) is computed by comparing \(\emb_t\) to the expected map embedding visible from \(x\):
\(
s_{\mathrm{sem}}(x_t) =
\cos\left(\emb_t, E_{\map}(x_t)\right)
\)
where \(E_{\map}(x_t)\) is obtained by lookup or ray-casting on the semantic map.

\section{\method}

\method{} has an offline semantic mapping stage and an online localization stage.

\begin{figure*}[t]
    \centering
    \safegraphics[width=1\textwidth]{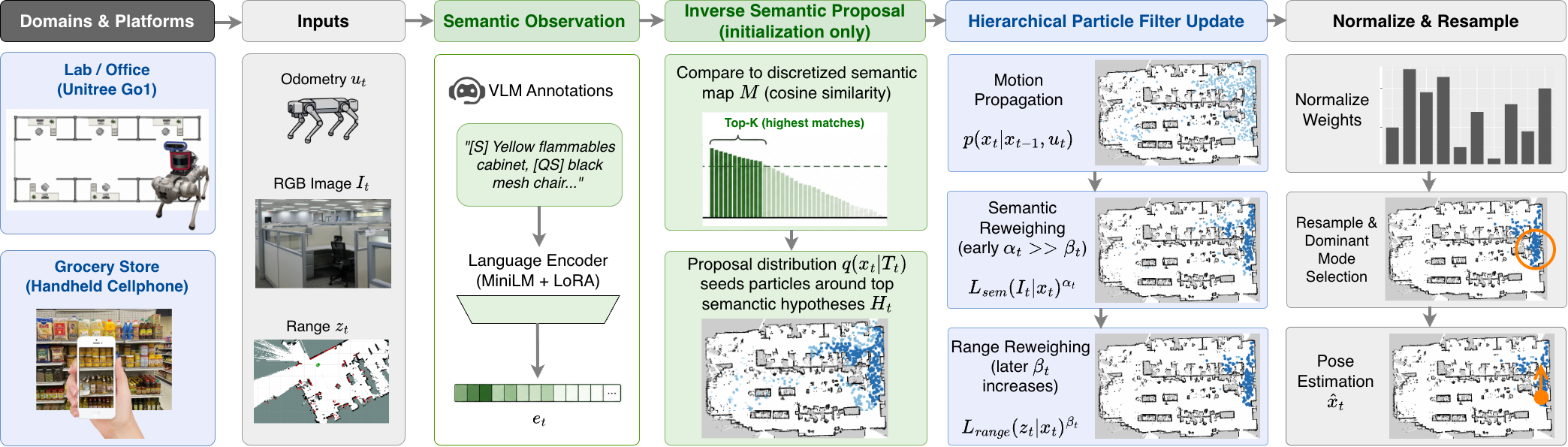}
    \captionsetup{font=small}
    \caption{\textbf{\method{} online localization pipeline.} Evaluated across two domains (quadruped \& cellphone), RGB images are VLM-annotated and encoded via a MiniLM that is LoRA-finetuned with quasi-static dropout augmentation for temporal robustness. Comparing this against the semantic map seeds initial particles via inverse proposal. A hierarchical filter then fuses odometry, semantics, and range data to predict the final pose.}
    \label{fig:system}
    \vspace{-1em}
\end{figure*}

\subsection{Offline Semantic Map Construction}

To build the offline map, \method{} first extracts objects from a mapping traversal. For each frame, it extracts bounding boxes in normalized coordinates and uses a VLM (\texttt{gemini-3-flash-preview}) to generate natural-language labels (5-10 words detailing color, material, object type, and visible text). The VLM also outputs a detection confidence score, following prior work on verbalized uncertainty estimation in VLMs~\citep{xuan2025seeing,xiong2024can,groot2024overconfidence} and a permanence classification: \texttt{[Static]} for structural elements, \texttt{[Quasi-Static]} for movable furniture, and \texttt{[Dynamic]} for transient objects. Low-confidence detections ($< 0.3$) and \texttt{[Dynamic]} items are filtered out.

Next, for each discretized free-space pose $p_i=(x_i,y_i,\theta_i)$, \method{} raycasts the camera field of view into the occupancy map up to a maximum range $r_{\max}$. All VLM-annotated objects visible from that pose are aggregated with their descriptions concatenated into a unified scene string and encoded via a sentence transformer $f_\phi$ (a LoRA fine-tuned MiniLM) to produce the embedding $\mathbf{e}_i \in \mathbb{R}^{384}$, yielding a discrete semantic map $\map_{\mathrm{text}} = \{(p_i, T_i, \emb_i)\}_{i=1}^{N}$.

The pose discretization differs by domain (Appendix~\ref{app:impl}). While the algorithmic interface is identical across domains, the low-level perception adapts to different platforms. In the lab domain, we use a Unitree Go1 quadruped (Figure~\ref{fig:go1}) with an Intel RealSense D455 and Velodyne VLP-16, relying on the VLM to segment and describe repetitive furniture/equipment. In the grocery domain, we use a consumer-grade smartphone and pre-propose shelf items using a YOLO-V9 detector~\citep{wang2024yolov9} fine-tuned on SKU-110K \citep{goldman2019precise}, using the VLM to extract product names, brands, and packaging details.

\begin{wrapfigure}[14]{r}{0.6\linewidth}
    \vspace{-1.2em}
    \centering
    \safegraphics[width=\linewidth , trim=0 0 0.6cm 0, clip]{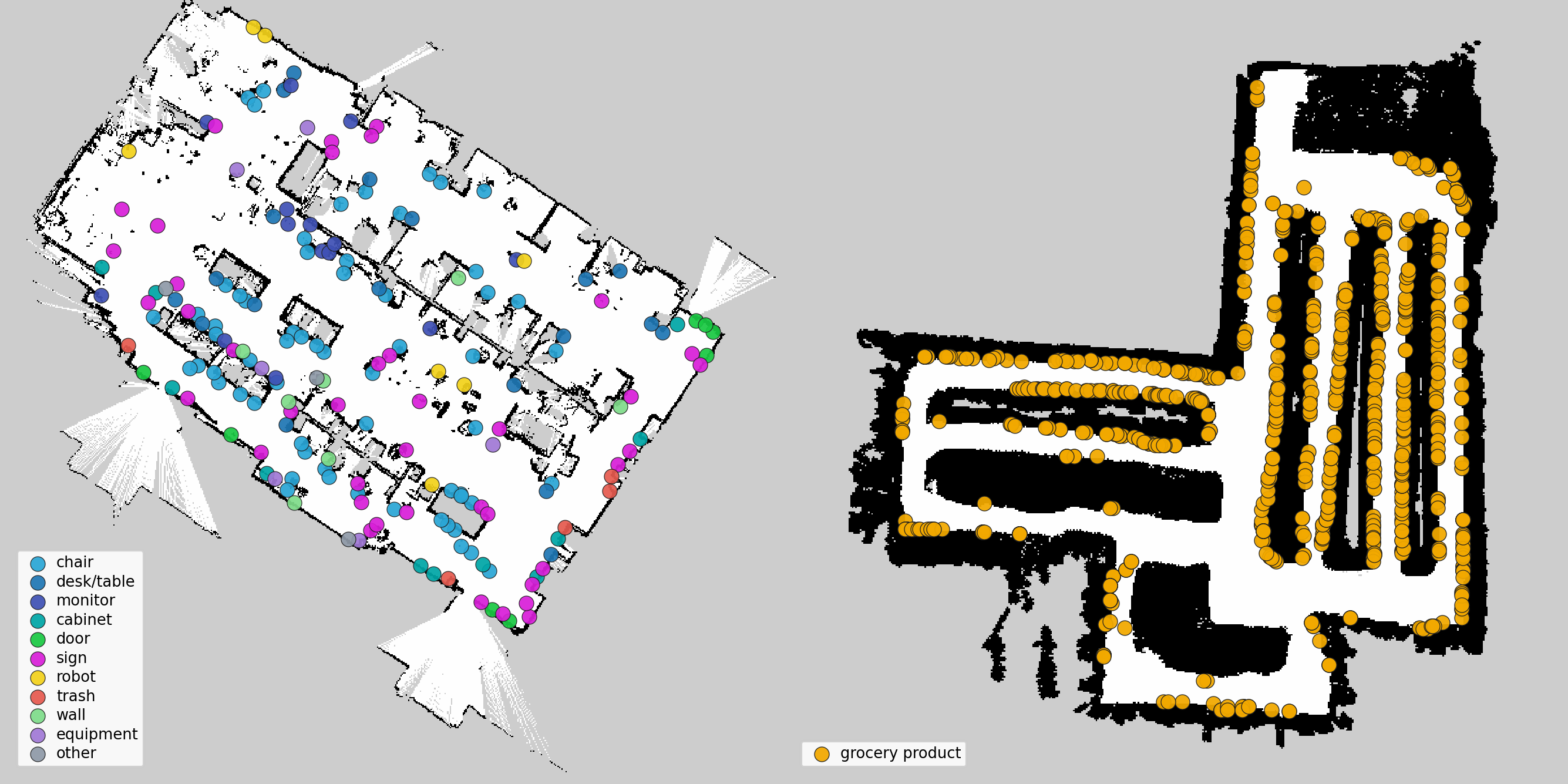}
    \captionsetup{font=small}
    \caption{\textbf{Semantic maps.} [Left] Lab map with repeated classes such as chairs, monitors, and desks. [Right] Grocery map with long-tail products.}
    \label{fig:semantic_maps}
\end{wrapfigure}

\subsection{Quasi-Static Data Augmentation}

The base sentence encoder is a frozen MiniLM-L6, a 22M-parameter model well-suited for resource-constrained computing platforms. We train a LoRA fine-tuned adapter using triplet loss on pose pairs generated from the offline semantic map. During the lab domain mapping, we queried the VLM twice per pose to obtain varied descriptions of the same scene, which served as positive pairs. Conversely, scenes separated by a distance greater than a threshold $T$ were utilized as negative pairs. For the grocery domain, we observed that the frozen MiniLM and the LoRA-finetuned model yield similar performance because the products naturally provide strong, distinct long-tail semantics. The lab, however, heavily benefits from fine-tuning because many regions share repeated object classes; the discriminative signal being subtle scene attributes and relative spatial arrangements.

To ensure the model remains robust against common environmental changes, we augment the positive text description pairs during fine-tuning using targeted dropouts. A quasi-static dropout simulates temporal change by removing transient items such as jackets, mugs, backpacks, and movable chairs, from the descriptions. These augmentations effectively simulate month-scale scene dynamics without requiring any physical rearrangement of the lab space.

\subsection{Online Localization}

Figure~\ref{fig:system} summarizes the online localization pipeline, detailed step-by-step in Appendix ~\ref{alg:vlmgloc}.

\noindent \textbf{VLM Observation Generation.} At test time, live semantic frames undergo the identical extraction, filtering, and concatenation process described in the offline mapping stage. The resulting scene string is passed through $f_\phi$ to produce the query embedding $\mathbf{e}_{\text{query}}$, ensuring the live observation and the prior map share a unified, highly discriminative feature space (e.g., precisely matching a ``yellow metal flammables storage cabinet'' rather than a generic ``cabinet'').

\noindent \textbf{Inverse Semantic Pose Proposal.}
The online algorithm begins with an inverse semantic observation proposal, which generates rich semantic pose candidates to quickly narrow the global search space. We compare the query embedding \(\emb_t\) against all map embeddings and retain the top \(K\) pose modes:
\(
\mathcal{H}_t =
\operatorname{TopK}_{p_i \in \map} \cos(\emb_t, \emb_i)
\)
These modes define a proposal distribution:
\(
q(\pose_t \mid I_t) =
\sum_{k=1}^{K} \pi_k
\mathcal{N}\!\left(\pose_t; \mu_k, \Sigma_k\right)
\)
where \(\mu_k\) is the pose of the retrieved map cell and \(\pi_k \propto \exp(s_k/\tau)\), with \(s_k=\cos(\emb_t,\emb_k)\). The temperature \(\tau\) controls how sharply particles concentrate around the highest-scoring semantic matches. This proposal step replaces a fraction of the initially uniform particles with samples drawn near the strongest semantic modes.

\noindent \textbf{Hierarchical Semantic MCL.}
After initialization, particles are propagated via odometry and reweighted using both semantic and range likelihoods:
\begin{align}
    \pose_t^{(n)} &\sim p(\pose_t | \pose_{t-1}^{(n)}, u_t), \\
    w_t^{(n)} &\propto w_{t-1}^{(n)}
        L_{\mathrm{sem}}(I_t | \pose_t^{(n)})^{\alpha_t}
        L_{\mathrm{range}}(z_t | \pose_t^{(n)})^{\beta_t}.
\end{align}

This factorization assumes that, conditioned on the robot pose and the map, the semantic image observation $I_t$ and the geometric range observation $z_t$ provide independent measurement evidence. 
The hierarchical nature of \method{} is controlled by scheduling the exponent weights \(\alpha_t\) and \(\beta_t\). During initial global localization, we enforce a strict semantic hierarchy ($\alpha_t \gg \beta_t$). The semantic likelihoods remain sharp while geometry is kept deliberately soft, preventing premature convergence to incorrect poses in geometrically aliased regions (e.g., identical aisles or workstations). Once the particle cloud has spatially converged, the range weight $\beta_t$ is increased. We define this spatial convergence as the moment the median particle spread \(\sigma_t = \operatorname{median}_{n} (\|\mathbf{x}_t^{(n)}-\operatorname{median}(\mathbf{x}_t^{(1:N)})\|_2)\) falls below a threshold \(\sigma_{\mathrm{conv}}=3.0\,\mathrm{m}\). This allows local geometry and odometry to precisely track the pose and reject incorrect modes.\\
The semantic likelihood is evaluated as temperature-scaled cosine similarity:
\(
    L_{\mathrm{sem}}(I_t | \pose_t^{(n)}) =
    \exp \left(\tau \cos(\emb_t, E_{\map}(\pose_t^{(n)}))\right).
\)
The geometric likelihood evaluates a 2D range observation (derived from the depth camera or lidar) against a distance transform of the occupied map cells. Finally, to select the best multi-modal hypotheses, \method{} estimates the final tracked pose by clustering high-weight particles via DBSCAN and returning the dominant weighted mode.

\section{Experimental Setup}
\vspace{-0.5em}
\begin{wrapfigure}[12]{r}[0pt]{0.32\textwidth}
    \vspace{-3.8em}
    \centering
    \safegraphics[
        width=\linewidth,
        trim=0 5cm 0 3cm,
        clip
    ]{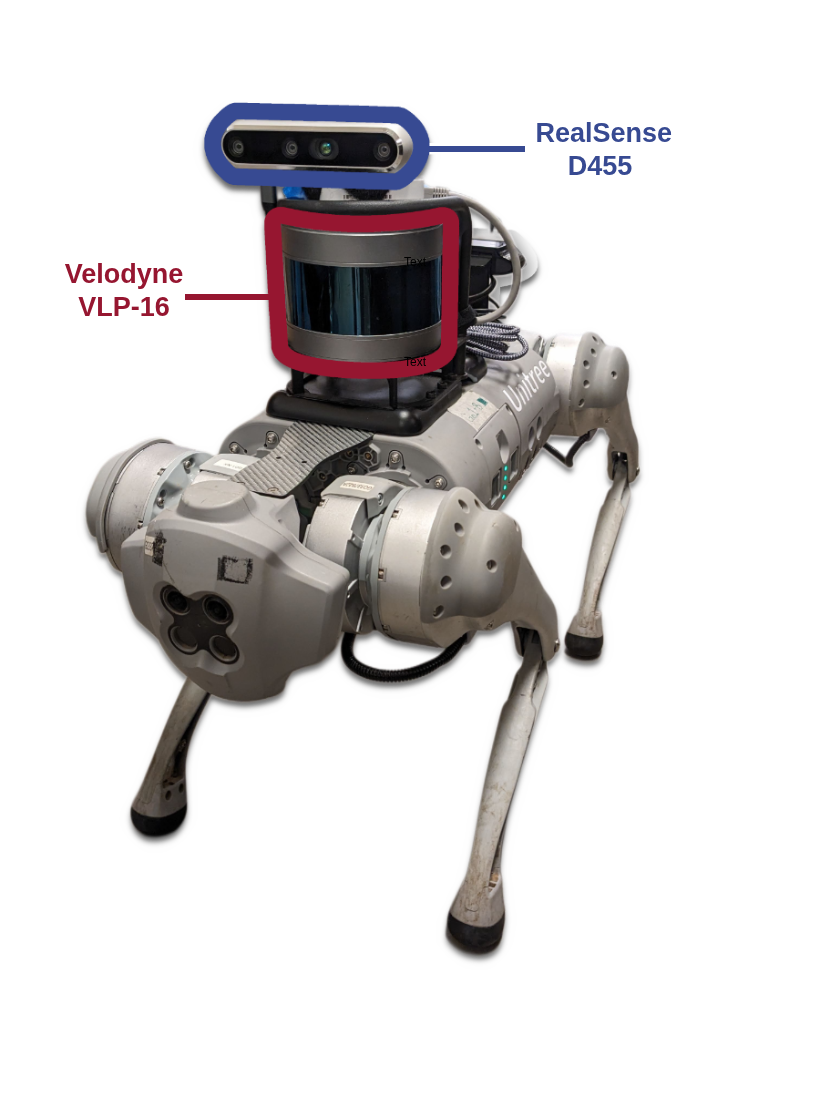}

    \captionsetup{
        width=0.95\linewidth,
        justification=raggedright,
        singlelinecheck=false,
        font=small
    }
    \caption{Go1 with RealSense D455 and Velodyne VLP-16.}
    \label{fig:go1}
\end{wrapfigure}
We collected 20 test trajectories per domain. To evaluate robustness against temporal changes, the lab domain trajectories we evaluate against were recorded one month after the initial mapping. During this period, several items, including bikes, whiteboards, robots, computer equipment, and chairs were moved or introduced, providing a realistic scenario for testing quasi-static resilience. Due to logistical constraints in the public retail space, the grocery trajectories were recorded later on the same day as the initial mapping. Key implementation parameters, including semantic encoder hyperparameters and particle filter configurations, are detailed in Appendix~\ref{app:impl}.

\textbf{Domain 1 (Lab):}
The 3,700 sq. ft. research facility is traversed by a Unitree Go1 quadruped (Figure~\ref{fig:go1}).
The environment is visually and geometrically repetitive, containing identical cubicles, desks, monitors, chairs, cabinets, and doors. Localization utilizes a SLAM Toolbox \citep{macenski2021slam} occupancy map paired with our discrete semantic map.

\textbf{Domain 2 (Grocery store):}
The grocery domain is a 3,500 sq. ft. international grocery store captured using a consumer-grade smartphone (RGB-D data). The store consists of parallel shelving units densely packed with long-tail products. The geometry is massively aliased, as multiple aisles produce nearly identical lidar range measurements. This means localization must rely entirely on the discriminative semantic fingerprints provided by product names, brands, and packaging.

\textbf{Ground Truth.} 
We utilize an offline map-aligned SLAM reference trajectory for both domains to ensure a high-fidelity baseline. We recorded exhaustive traversals achieving full map coverage. Because robust Iterative Closest Point (ICP)~\citep{besl1992method} alignment requires this extensive spatial overlap, we use it strictly offline to align these rebuilt environment maps against the prior mapping traversals, obtaining highly accurate translation and rotation matrices. We then chunked these aligned traversals with random starting points into the individual test trajectories used for evaluation.

\noindent\textbf{Baselines.}
\textbf{AMCL:}
The AMCL baseline uses the same particle-filter infrastructure but without semantic observations. It uses range likelihoods and odometry only.



\textbf{FCS-MCL (Fixed-Class Semantic MCL):}
To evaluate fixed-class semantic localization, we implemented the ShelfAware \citep{agrawal2026shelfaware} baseline (tailored for the grocery domain), which alongside \citet{zimmermanLongTermLocalizationUsing2023} represents the standard semantic paradigm. This categorical particle filter captures the spatial distribution of these classes (using predefined category counts, mean distances, and mean relative angles) to model high-level semantic layouts. Adapting a domain-specific vision front-end, the grocery state vector tracks 20 product categories (e.g., rice, lentils, spices). For the lab, it uses a vocabulary like chair, desk, monitor, sign, cabinet, door, and robot.

\textbf{Metrics:} Following established conventions in the literature \citep{zimmermanLongTermLocalizationUsing2023, zimmerman2022robust, agrawal2026shelfaware}, we define \textbf{convergence} as the estimated pose falling within $0.7$~m translation and $\pi/4$~rad rotation of the ground truth. To decouple initial pose estimation from continuous state tracking, we evaluate \method{} via:

\noindent \textbf{Global localization success:} A trial is successful if initial convergence occurs within the first 95\% of the trajectory.\\
\textbf{Global localization error (G-error):} The absolute translation and rotation error measured at the initial convergence, capturing the accuracy of the system the moment global ambiguity is resolved.\\
\textbf{Whole Trajectory ATE (W-ATE):} The Absolute Trajectory Error (translation and rotation RMSE) computed from the moment of initial global convergence until the absolute end of the sequence.\\
\textbf{Stable tracking success:} A trial is successful if it contains a final tracking that begins before the 95\% cutoff and remains strictly within the convergence threshold until the end of the sequence.\\
\textbf{Stable Tracking ATE (T-ATE):} The Absolute Trajectory Error (translation and rotation RMSE) computed over the successfully tracked trajectory.

\section{Results}

\textbf{Hypothesis 1: Rich Text Features Resolve Aliasing:}
We first demonstrate that rich semantics resolve severe geometric aliasing better than both pure geometry and top-level class vectors across two distinct domains.
We run each trajectory 5 times to account for stochasticity and report the mean. Table~\ref{tab:lab_unseen_baselines} reports the unseen lab data comparison against AMCL and FCS-MCL on the exact same trajectory chunks. AMCL succeeds on only 4\% of the runs due to repeating cubicles. FCS-MCL succeeds on 10\%; its class vector erases the specific details (material, text, color, state) that distinguish repeated lab objects. In contrast, \method{} reaches 74\% global localization success. The tracking success numbers remain lower than global localization success. However, the lab tracking trajectories are often visually close even when some frames fail the threshold. When we evaluate tracking using an ATE threshold, meaning small trajectory violations are allowed as long as the mean error remains under the threshold, the stable tracking success improves from 46\% to 60\%.

\begin{table}[h]
\centering
\caption{Lab evaluation after one month of mapping. Metrics include Global Localization (Global \%) and Tracking Success (Tracking \%). All errors are reported as translation (m) / rotation (rad).}
\label{tab:lab_unseen_baselines}
\small
\begin{tabular}{lccccc}
\toprule
Method & Global \% & G-error (t/r) & W-ATE (t/r) & Tracking \% & T-ATE (t/r) \\
\midrule
AMCL & 4\%  & \textbf{0.45 / 0.12} & 6.72 / 1.44 & 0\% & n/a \\
FCS-MCL & 10\% & 0.50 / 0.32 & 0.79 / 0.29 & 0\% & n/a \\
\method{} & \textbf{74\%} & 0.51 / 0.23 & \textbf{0.66 / 0.20} & \textbf{46\%} & \textbf{0.41 / 0.13} \\
\bottomrule
\end{tabular}
\end{table}

Table~\ref{tab:grocery} shows similar findings in the grocery domain. Even against a competitive class-vector baseline utilizing 20 grocery categories, \method{} improves global localization success from 30\% to 70\%. While the depth profiles of parallel aisles are identical, the long-tail language of product brands, packaging, and ingredient names provides a unique semantic fingerprint that allows the filter to converge reliably. While baselines occasionally report marginally lower tracking errors (T-ATE), this can be attributed to the fact that methods like AMCL and FCS-MCL succeed far less often. In contrast, \method{} maintains comparable RMSEs while successfully localizing on a vastly larger set, demonstrating a significant contribution to system reliability. Sensitivity analysis (Appendix~\ref{app:sensitivity}) reveals that minor relaxations of the spatial convergence thresholds boost \method{}'s global localization success to 95\%. Isolating translation error accounts for the mirrored semantic viewpoints prevalent in the grocery domain, yielding significant gains not observed in baseline methods.

\textbf{Hypothesis 2: Implicit Quality Filtering:}
We hypothesize that VLMs act as cognitive filters, suppressing noisy visual evidence. To validate this, we conducted an ablation in the grocery domain where blurry or unreadable product crops were intentionally retained in both the map and the query text. These low-quality images are particularly prevalent in this domain due to the handheld cellphone capture. As shown in Table~\ref{tab:ablations}, forcing the system to utilize these noisy observations degrades global localization by a 29\% relative drop. Retaining unreadable items makes the semantics noisier, delays convergence, and prevents the filter from anchoring to reliable landmarks.

\textbf{Hypothesis 3: Permanence Reasoning \& Training Data Generation via Augmentation:}
Finally, we demonstrate within the lab domain that \method{} can characterize transient features to improve localization. As highlighted in Table~\ref{tab:ablations}, relying on a frozen MiniLM encoder results in an 18.75\% relative drop in global localization success compared to our LoRA finetuned adapter, which is explicitly trained with targeted text-dropout augmentation that removes quasi-static items (jackets, backpacks, movable chairs). Furthermore, entirely removing VLM-generated permanence and quality tags from the descriptions causes a measurable 8\% relative drop in success, proving that reasoning over these cues yields richer semantics crucial for robust localization.

\vspace{-0.8em}
\begin{table}[h]
\centering
\caption{Grocery store evaluation over 20 trajectories in a 3,500 sq. ft. space.}
\label{tab:grocery}
\small
\begin{tabular}{lccccc}
\toprule
Method & Global \% & G-error (t/r) & W-ATE (t/r) & Track \% & T-ATE (t/r) \\
\midrule
AMCL & 20\%  & 0.50 / \textbf{0.10} & 2.08 / 0.59 & 5\% & 0.31 / \textbf{0.15} \\
FCS-MCL & 30\%  & \textbf{0.30} / 0.26 & 0.82 / \textbf{0.39} & 25\% & \textbf{0.30} / 0.16 \\
\method{} & \textbf{70\%} & 0.32 / 0.31 & \textbf{0.77} / 0.64 & \textbf{65\%} & 0.31 / 0.18 \\
\bottomrule
\end{tabular}
\vspace{-1em}
\end{table}


\vspace{-0.5em}
\begin{table}[H]
\centering
\caption{Ablation impacts on global localization success across domains.}
\label{tab:ablations}
\small
\begin{tabular}{lc}
\toprule
Condition & Global \% $\Delta$ \\
\midrule
Retained blurry crops (Quality Filtering) & -29\% (grocery)\\
Removed permanence tags (Permanence Reasoning) & -8\% (lab)\\
LoRA with Augmentation vs. Frozen MiniLM (No Fine-tuning)& -18.75\% (lab)\\
\bottomrule
\end{tabular}
\vspace{-1em}
\end{table}

\section{Qualitative Analysis}
\begin{figure}[!htbp]
    \centering
    \safegraphics[width=0.9\textwidth]{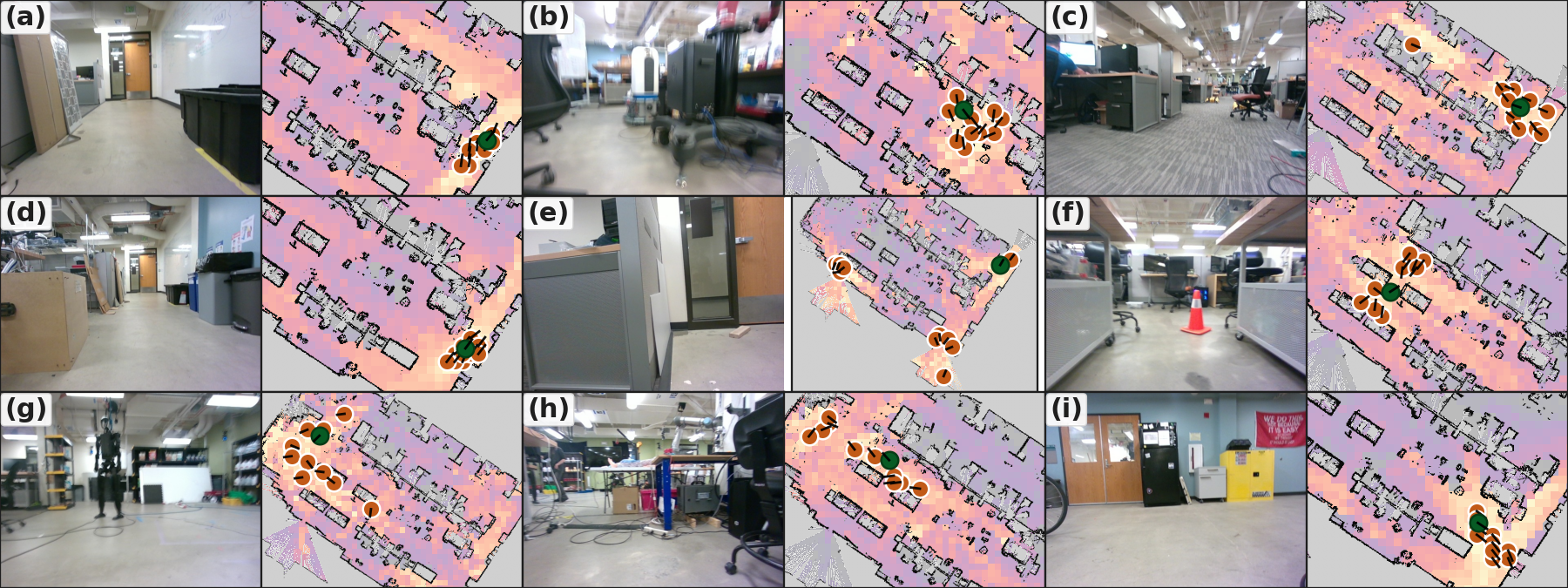}
    \captionsetup{font=small}
    \caption{\textbf{Top-10 inverse semantic pose proposals.} For each live RGB observation (left), orange markers denote the top-10 retrieved semantic pose hypotheses within the prior map (right), while the green marker indicates the ground truth pose. The underlying heatmap visualizes the dense semantic cosine similarity, with lighter (yellow) regions representing higher retrieval scores. The proposal mechanism successfully anchors the robot despite severe motion blur (b), geometrically aliased or mirrored viewpoints (c, f, g, h), repeated-door ambiguity (e), and month-scale scene changes such as relocated objects or new clutter (f, g).}
    \label{fig:topk}
    \vspace{-1em}
\end{figure}

\begin{figure}[!htbp]
    \centering
    \safegraphics[width=0.92\linewidth]{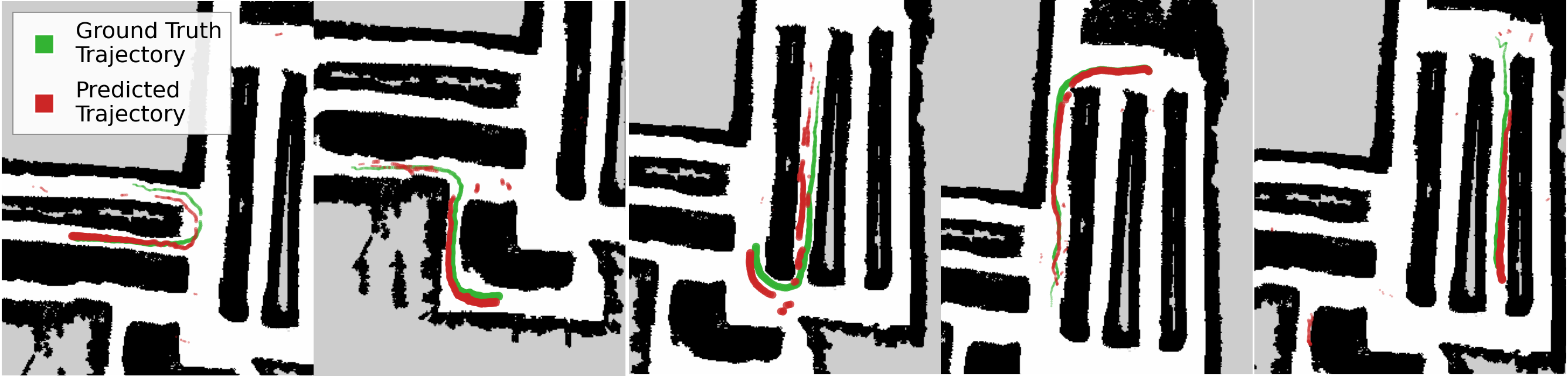}
    \captionsetup{font=small}
    \caption{Example grocery trajectories with ground truth and \method{} pose
    estimates. Thinner to thicker correspond to the temporal progression of the trajectory.}
    \label{fig:grocery_trajs}
    \vspace{-1em}
\end{figure}

\vspace{-0.5em}
\textbf{Inverse Semantic Proposal.} As shown in Figure~\ref{fig:topk}, inverse semantic retrieval provides strong global pose estimates before temporal particle filtering. The qualitative results highlight the resilience of our semantic proposal against common real-world failure modes. For instance, while mirrored viewpoints induce some local orientation scatter, the semantic hypotheses still concentrate in the correct global region. Furthermore, under severe perceptual ambiguity, such as featureless areas with identical doors, the system degrades gracefully, proposing multiple plausible regions. \method{} handles month-scale quasi-static changes well. Although items shifted between mapping and testing, the unified scene descriptions maintain enough contextual overlap to retrieve the correct area, showing applicability to long-term deployment in active spaces. \\
\textbf{Trajectory Examples.} As a counterpart in the retail domain, Figure~\ref{fig:grocery_trajs} illustrates example grocery trajectories with ground truth and \method{} pose estimates. In this domain, semantic convergence and tracking are usually aligned because product identities sharply disambiguate repeated aisles.

\vspace{-0.5em}
\section{Discussion}
\vspace{-0.5em}

\textbf{VLMs provide more than object names.}
The primary value of the VLM front end lies beyond simple object categorization. It preserves fine-grained, open-vocabulary evidence: colors, materials, readable text, packaging, and object state. It also filters low-quality evidence by characterizing noisy observations. This is why \method{} outperforms class-vector semantic baselines even though both rely on ``semantics.'' For example, a standard detector label such as `cabinet' loses warning text and material context; `chair' loses the mesh back and orange cushion details; and `sign' loses the actual text. Appendix \ref{app:examples} provides high-similarity query/map description pairs demonstrating how this fine-grained evidence yields strong map retrieval scores. 

\textbf{Domain-general interface, domain-specific front ends.}
Both domains produce the exact same interface to the localizer: structured text and normalized embeddings. While previous baselines require a different class vocabulary and scoring vector engineered specifically for each domain, \method{} removes the need for a dedicated vision front end for each domain.

\textbf{Hardware Accessibility and Form Factor.} The grocery evaluation demonstrates that robust localization can be achieved using a standard cellphone as the primary sensor suite. This lowers the hardware barrier to entry, making the framework applicable for lightweight mobile robots and cellphone-based assistive navigation technologies.

\textbf{Runtime and Practical Deployment.}
While the cached particle filter runs at 8 Hz (with Go1 sensors) and 44 Hz (with cellphone sensors), live synchronous cloud-VLM API calls yield an effective online rate of roughly 0.21 Hz for the lab and 0.91 Hz for the grocery store (using a semantic update interval of 5 frames, which was standard across our experiments). This sparse update rate is practical for resolving global ambiguity. During tracking, \method{} fuses these intermittent semantic observations with high-frequency geometric filtering and odometry to maintain smooth pose estimates.
Full latency profiling is detailed in Appendix \ref{app:runtime}. Despite the latency bottleneck introduced by cloud APIs, the resulting accuracy and unified semantic front-end make combining VLMs within a probabilistic framework highly promising. Notably, while the grocery pipeline still utilizes a generic product class detector at the front, \method{} entirely removes the need for domain-specific clustering, spatial aggregation rules, and dense feature extractors required by prior systems.

\textbf{Limitations:}
Our method inherently relies on rich semantic features to disambiguate scenes, meaning performance degrades in visually barren environments. Furthermore, balancing particle convergence for smooth pose estimation against the need to explore new hypotheses remains challenging. When likelihoods drop, it is difficult to determine whether the cause is physical scene changes, a lack of local features, or if particles have drifted into an incorrect mode requiring global re-exploration. Finally, while the current cloud-based VLM latency restricts high-frequency semantic tracking, rapid advancements in lightweight, on-device VLMs promise to alleviate this constraint.

\textbf{Conclusion:}
\method{} reframes VLMs as semantic observation generators for probabilistic robot localization, replacing brittle, domain-specific object detectors with a unified natural language interface. By converting query images into language-rich pose proposals and hierarchical semantic likelihoods, the system resolves severe global ambiguity in environments where geometry alone fails. Across vastly different domains: from repetitive laboratory cubicles navigated by a quadruped to dense retail aisles mapped via a consumer cellphone, the identical semantic MCL framework achieves robust global localization. Ultimately, \method{} demonstrates that rich open-vocabulary descriptions, implicit quality filtering, and targeted quasi-static reasoning offer a highly portable, hardware-agnostic solution for autonomous indoor navigation.

\bibliography{main}
\clearpage

\appendix

\section*{Appendices}

\section{Implementation and Reproducibility Details}
\label{app:impl}

Table~\ref{tab:implementation} records the main parameters for the headline
runs. 

\begin{table}[H]
\centering
\caption{Key implementation parameters.}
\label{tab:implementation}
\small
\begin{tabularx}{\linewidth}{lX}
\toprule
Component & Value \\
\midrule
Compute & Dell G15 laptop (Intel Core i7-11800H (2.3 GHz); NVIDIA RTX~3060, 6GB VRAM; 32GB RAM)\\
Lab semantic map & SLAM Toolbox occupancy map at $0.05$ m/px; semantic grid spacing $0.50$ m; 24 heading bins; $80.5^\circ$ camera FOV; 10 m raycast; objects retained after $\geq 3$ mapping observations \\
Grocery semantic map & Occupancy map at $0.05$ m/px; semantic grid spacing $0.50$ m; 8 heading bins; $60^\circ$ camera FOV; 6 m raycast \\
Semantic encoder & MiniLM-L6-v2 text encoder with 384-d embeddings; lab uses triplet LoRA; grocery uses the frozen encoder \\
Lab LoRA training & Triplet loss, margin 0.5; LoRA rank 16, alpha 32, module dropout 0.1; 30 epoch schedule; batch size 32; learning rate $2\times10^{-5}$; $T=3m$ \\
Lab text augmentation & Positive-pair augmentation with quasi-static dropout, dynamic removal, item shuffle, subset-drop, composed perturbations, and exhaustive quasi-static variants; quasi-static and subset dropout remove 1--2 items \\
Lab localization & Semantic update every 5 frame; top-400 semantic proposal; candidate temperature 10; semantic seed noise $\sigma_{xy}=0.30$ m, $\sigma_\theta=0.16$ rad; DBSCAN pose estimate with $\epsilon=1.0$ m; lidar weight 0.10 \\
Grocery localization & Semantic update every 5 frames;  semantic temperature 10; adaptive depth/lidar cap 0.50; base lidar weight 0.10; DBSCAN pose estimate with $\epsilon=1.0$ m \\
Semantic/range weights &
Post-convergence schedule with \(\alpha_t=1\). Before spatial convergence,
\(\beta_t=0\); after \(q_{50}<3.0\,\mathrm{m}\), \(\beta_t=0.10\), where
\(q_{50}\) is the median particle distance from the particle-cloud spatial median.
For the lab headline setting, regular online range reweighting is disabled
(\(\beta_t=0\)); lidar is used instead for semantic-candidate reranking with
\(\beta_{\mathrm{rerank}}=1.0\) and top-\(k=10\) lidar pose estimation. \\

Particle filter & KLD-adaptive MCL with 1500-particle cap and 200-particle floor; motion noise $(0.05, 0.02, 0.02, 0.01)$; laser $\sigma_{\mathrm{hit}}=0.30$ m \\
VLM & \texttt{gemini-3-flash-preview} (Temp: 0.3) \\
Pose estimate & Dominant DBSCAN-weighted particle mode over position and circular heading \\
\bottomrule
\end{tabularx}
\end{table}

\section{Runtime and Latency Profiling}
\label{app:runtime}

\begin{table}[H]
\centering
\caption{Runtime summary. Cached particle filter (PF) Hz excludes live VLM annotation. Live annotation uses a semantic interval of 5 for both domains to estimate practical online Hz.}
\label{tab:runtime}
\small
\begin{tabular}{lccc}
\toprule
Domain & Cached PF Hz & Live VLM Mean Latency & Est. Online Hz \\
\midrule
Lab & 8.64 & 23.44 s & 0.21 \\
Grocery & 44.27 & 5.36 s & 0.91 \\
\bottomrule
\end{tabular}
\end{table}

\section{Dominant-Mode Pose Estimate}
\label{app:dbscan}
The posterior of the particle filter can remain multi-modal. We therefore do not estimate pose by a global weighted mean over all particles. Instead, we cluster high-weight particles with DBSCAN over position and circular heading features and return the dominant weighted mode. This better matches the localization problem: the robot should report one coherent pose hypothesis, not the average of two different aisles or two mirrored lab regions.

\clearpage

\section{Online Localization Algorithm}
\label{alg:vlmgloc}
\begin{algorithm}[H]
\caption{\method{} Online Localization}
\small
\begin{algorithmic}[1]
\Require semantic map \(\map\), particles \(\{\pose^{(n)}, w^{(n)}\}\), 
\Statex \hspace{1.2em} odometry \(u_t\), image \(I_t\), range \(z_t\)
\State Propagate particles with odometry motion model \(p(\pose_t|\pose_{t-1},u_t)\)
\If{semantic frame is available}
    \State \(T_t \gets \mathrm{VLMAnnotation}(I_t)\)
    \State \(\emb_t \gets \mathrm{TextEncoder}(T_t)\)
    \If{not initialized}
        \State \(\mathcal{H}_t \gets \operatorname{TopK}_{p_i\in\map}\cos(\emb_t,\emb_i)\)
        \State Seed a fraction of particles from \(q(\pose_t|I_t)\) around \(\mathcal{H}_t\)
    \EndIf
    \State Reweight particles by \(L_{\mathrm{sem}}(I_t|\pose_t)^{\alpha_t}\)
\EndIf
\If{range observation is available}
    \State Reweight or rerank particles by \(L_{\mathrm{range}}(z_t|\pose_t)^{\beta_t}\)
\EndIf
\State Normalize weights
\State Resample when effective particle count is low
\State \(\hat{\pose}_t \gets\) dominant DBSCAN weighted mode
\State \Return \(\hat{\pose}_t\)
\end{algorithmic}
\end{algorithm}

\section{Convergence Sensitivity Analysis}
\label{app:sensitivity}

As discussed in Section 6, the strict $\pi/4$ rotation threshold occasionally penalizes valid localizations where particles observe the exact same semantic and range targets from mirrored viewpoints (a common occurrence in parallel grocery aisles). Table~\ref{tab:grocery_thresholds} demonstrates that when evaluating purely on translation, or when slightly relaxing the strict spatial threshold, \method{}'s success rate climbs significantly, whereas baselines fail to see similar improvements. 

\begin{table}[H]
\centering
\caption{Sensitivity of grocery store localization and tracking success to relaxed spatial thresholds (0.7 m / 0.8 m / 0.9 m). The heading threshold is fixed at $\pi/4$~rad for Joint metrics. Table~\ref{tab:grocery} shows results corresponding to the highlighted numbers and threshold.}
\label{tab:grocery_thresholds}
\small
\begin{tabular}{lcccc}
\toprule
\multirow{2}{*}{Method} & \multicolumn{2}{c}{Translation-Only (\%)} & \multicolumn{2}{c}{Joint Translation + Rotation (\%)} \\
\cmidrule(lr){2-3} \cmidrule(lr){4-5}
& Global Success \% & Tracking Success \% & Global Success \% & Tracking Success \% \\
\midrule
AMCL & 30 / 30 / 30 & 10 / 15 / 15 & 20 / 20 / 20 & 5 / 10 / 10 \\
FCS-MCL & 35 / 35 / 35 & 20 / 30 / 30 & 30 / 30 / 30 & 25 / 25 / 25 \\
\method{} & \textbf{85 / 90 / 95} & \textbf{70 / 75 / 80} & \textbf{\colorbox{pink!50}{70} / 75 / 80} & \textbf{\colorbox{pink!50}{65} / 70 / 70} \\
\bottomrule
\end{tabular}
\end{table}

\section{Ground Truth Acquisition Details}
\label{app:ground_truth}

To establish ground truth, we initially evaluated a ceiling-mounted AprilTag constellation in the lab domain. However, disagreements among simultaneously visible tags produced systematic pose uncertainties on the order of 0.75~m, which is too high for rigorous evaluation. Furthermore, deploying such physical infrastructure in the public grocery store was logistically infeasible. Therefore, we utilized the offline map-aligned SLAM reference trajectories described in Section 5, which provided a highly accurate and consistent baseline across both domains without requiring physical environmental modifications.

\section{Rich VLM Description Examples}
\begin{table}[H]
\centering
\caption{Qualitative comparison between generic object classes and \method{}'s tagged, open-vocabulary descriptions. \texttt{[Dynamic] [Blurry]} objects are filtered before map/query embedding so they are not present in the following.}
\label{tab:vlm_vs_class}
\small
\begin{tabularx}{\textwidth}{llX}
\toprule
\textbf{Domain} & \textbf{Generic Object Class} & \textbf{VLM Rich Description} \\
\midrule
Lab & Chair &
\texttt{[Quasi-Static]} black mesh office chair with orange fabric seat cushion and wheels. \\

Lab & Cabinet &
\texttt{[Quasi-Static]} yellow metal \texttt{JUSTRITE} flammables storage cabinet with warning labels. \\

Lab & Door &
\texttt{[Static]} light-brown wooden door with rectangular glass panel, silver handle, and metal kick plate. \\

Lab & Whiteboard &
\texttt{[Static]} large whiteboard with checkerboard AprilTag markers and handwritten \texttt{VLA} notes. \\

Lab & Robot &
\texttt{[Quasi-Static]} white-blue mobile robot base beside a red robotic arm, blue cables. \\

Grocery & Lentils &
Swad Yellow Vatana (Peas), Swad, Lentils. \\

Grocery & Condiment &
Mother's Recipe Mango Pickle, Mother's Recipe, Condiments. \\

Grocery & Oil &
Jiva Organic Mustard Oil, Jiva, Oil. \\

Grocery & Spice &
MDH Pav Bhaji Masala, MDH, Spices. \\

\bottomrule
\end{tabularx}
\end{table}

\clearpage
\section{High-Similarity Description Pairs}
\label{app:examples}

\begin{table}[H]
\centering
\caption{VLM description pairs from held-out lab queries and the semantic text map. The top rows are good semantic matches; rows below the separator are weaker/confusable pairs that share generic object classes but differ in fine-grained attributes or scene context. Similarity is cosine score from the lab LoRA encoder. For grocery product descriptions, blurry/unreadable and dynamic detections are filtered before embedding.}
\label{tab:description_examples}
\small
\begin{tabularx}{\textwidth}{p{0.42\textwidth} p{0.42\textwidth} c}
\toprule
Query description excerpt & Raycasted map description excerpt & Sim. \\
\midrule
\texttt{[Quasi-Static]} red fabric banner with white text \texttt{WE DO THIS NOT BECAUSE IT IS EASY}; \texttt{[Quasi-Static]} yellow metal \hlcabinet{cabinet} with \texttt{JUSTRITE} and \texttt{FLAMMABLES} labels; gray rolling \hlcabinet{cabinet}. &
\texttt{[Quasi-Static]} red fabric banner with matching white text; \texttt{[Quasi-Static]} yellow safety \hlcabinet{cabinet} with \texttt{FLAMMABLES KEEP FIRE AWAY}; nearby wall signs and wooden \hldoor{door}. &
0.813 \\

\texttt{[Static]} brown wooden double \hldoor{doors} with glass panels; \texttt{[Quasi-Static]} black refrigerator with papers; \texttt{[Quasi-Static]} white emergency \hlcabinet{cabinet}; yellow \hlcabinet{cabinet}. &
\texttt{[Quasi-Static]} yellow flammables storage \hlcabinet{cabinet} with warning labels; red banner; \texttt{[Static]} light-brown \hldoor{door} with metal kick plate; blue wall fixtures. &
0.782 \\

\texttt{[Quasi-Static]} tan plywood crate with latch; silver wire shelving; cardboard sheets; gray industrial \hlcabinet{cabinet}; \texttt{[Static]} light-brown \hldoor{door}. &
\texttt{[Static]} light-brown \hldoor{door} with glass window; \texttt{[Quasi-Static]} gray storage \hlcabinet{cabinet}; blue recycling bin; black storage bin; whiteboard/storage area. &
0.768 \\

\midrule
\multicolumn{3}{c}{\footnotesize Weaker pairs sharing generic object classes} \\
\midrule

\texttt{[Static]} light-brown wood-grain \hldoor{door} with silver kick plate; dark metal \hldoor{door} frame; \texttt{[Quasi-Static]} cubicle partition and gray storage \hlcabinet{cabinet}. &
\texttt{[Static]} brown wooden \hldoor{door} with blue-white sign; \texttt{[Static]} orange wooden \hldoor{door}; \texttt{[Static]} fire-extinguisher \hlcabinet{cabinet}; repeated wall boards. &
0.605 \\

\texttt{[Quasi-Static]} wood and metal \hlbench{workbench}; gray storage \hlcabinet{cabinet}; red/yellow storage bins; black fabric office \hlchair{chair}; backpack; fire-extinguisher \hlcabinet{cabinet}. &
\texttt{[Quasi-Static]} black monitor on a \hlbench{workbench}; computer tower; gray storage \hlcabinet{cabinets}; black padded rolling \hlchair{stool}; yellow storage bin. &
0.526 \\

\texttt{[Quasi-Static]} large monitor on desk; gray cubicle desk; blue \hlchair{chair}; black computer tower; cardboard boxes; trash bin. &
\texttt{[Quasi-Static]} black mesh \hlchair{chair} with orange seat; gray storage \hlcabinet{cabinet}; black framed \hlmonitor{monitor}; wood desk edge; repeated workstation objects. &
0.430 \\
\bottomrule
\end{tabularx}
\end{table}

\section{Gemini Annotation Prompts}
\label{app:gemini_prompts}

\begin{figure*}[t]
\centering
\begin{minipage}{0.95\textwidth}
\footnotesize
\textbf{Lab / office object annotation prompt.}
\begin{verbatim}
Detect all distinct objects in this lab/office image. For each object, provide:
1. A bounding box [ymin, xmin, ymax, xmax] in normalized 0-1000 coordinates
2. A SHORT label (5-15 words max): color + material + object type. Include any readable text.
3. Permanence: [S]=structural, [QS]=quasi-static furniture/equipment, [D]=dynamic people
4. Confidence (0.0-1.0): how certain you are this is a real, distinct object 
(not blur/shadow/artifact)

Return as JSON array:
[{"box": [y1, x1, y2, x2], "label": "brief identity", "perm": "S/QS/D", "conf": 0.9}]

Rules:
- Do NOT include position in image (left, right, foreground)
- Do NOT include spatial relationships to other objects
- Do NOT detect plain floor or plain ceiling
- DO detect: walls with signs/text, distinctive fixtures, furniture, equipment, signs, monitors
- Be SPECIFIC: not "chair" but "black mesh office chair with orange cushion"
- Read and include any visible TEXT on signs, labels, room numbers
- Ignore motion blur artifacts and infrared dot patterns; only detect clearly visible objects
- conf should be LOW for blurry/unclear objects
\end{verbatim}
\vspace{0.75em}

\textbf{Grocery product-crop annotation prompt.}
\begin{verbatim}
You are analyzing a grid of {n} product images detected on shelves in an Indian grocery store.
Each image is labeled with an ID (C0, C1, etc.) and is a crop from a shelf detector.

For EACH product crop, identify it and provide:
{
  "id": "C0",
  "product_name": "specific product name (brand + product if readable)",
  "brand": "brand name if visible, else null",
  "category": "e.g., Spices, Lentils, Beverages, Snacks, Oil, Tea, Rice, Flour, Condiments, 
  Sweets, Frozen, Dairy, Personal Care, Household",
  "is_blurry": true or false
}

Rules:
- Be specific: say "MDH Chana Masala" not just "spice". Say "Laxmi Moong Dal" not just "lentils".
- If the crop is too blurry to identify the product, set is_blurry to true 
and still fill your best guess.
- If there are NO identifiable products (e.g., empty shelf, wall, floor, ceiling), return:
  {"products": [], "suitable_for_localization": false}
- Otherwise set "suitable_for_localization": true

Return ONLY valid JSON with a "products" array and "suitable_for_localization" boolean.
No markdown fences.
\end{verbatim}
\end{minipage}
\caption{Gemini prompts used for VLM annotation in the lab and grocery domains. The lab prompt operates on full RGB frames and requests object boxes, rich labels, permanence tags, and confidence. The grocery prompt operates on YOLO-detected product crop grids and requests product identity, category, and blur filtering metadata.}
\label{fig:gemini_prompts}
\end{figure*}

\end{document}